# SMILE: A Super-resolution Guided Multi-task Learning Method for Hyperspectral Unmixing

Ruiying Li, Bin Pan, Qiaoying Qu, Xia Xu and Zhenwei Shi

*Abstract*—The performance of hyperspectral unmixing may be constrained by low spatial resolution, which can be enhanced using super-resolution in a multitask learning way. However, integrating super-resolution and unmixing directly may suffer two challenges: Task affinity is not verified, and the convergence of unmixing is not guaranteed. To address the above issues, in this paper, we provide theoretical analysis and propose super- resolution guided multi-task learning method for hyperspectral unmixing (SMILE). The provided theoretical analysis validates feasibility of multitask learning way and verifies task affinity, which consists of relationship and existence theorems by proving the positive guidance of super-resolution. The proposed frame- work generalizes positive information from super-resolution to unmixing by learning both shared and specific representations. Moreover, to guarantee the convergence, we provide the acces- sibility theorem by proving the optimal solution of unmixing. The major contributions of SMILE include providing progressive theoretical support, and designing a new framework for unmixing under the guidance of super-resolution. Our experiments on both synthetic and real datasets have substantiate the usefulness of our work. The code is available online[1].

## I. INTRODUCTION

HYPERSPECTRAL images (HSIs) provide more subtle and valuable spectral information than common images [1]–[3]. This advantage of HSIs leads to meaningful attentions in some fields, such as mineral resource exploitation, military surveillance and forest resource survey [4]–[6]. However, there is an inevitable trade-off in HSIs between high spectral resolution and low spatial resolution, which leads to the existence of mixed pixels [7]. This problem of mixture can be addressed by hyperspectral unmixing (HU) techniques that divide pixels into the spectra of pure materials (*endmembers*) and their corresponding fractions (*abundances*) [8]. Different mixing models on which current unmixing algorithms depend can be classified into linear mixing model (LMM) and nonlinear mixing model [9]. LMM treats mixed pixels as linear weighted combinations of pure materials, with their abundance as corresponding weight coefficients [10], [11].

Due to the effectiveness and simplicity of LMM, it has attracted significant interest from researchers, making it a widely utilized approach in unmixing [12], [13]. Algorithms based on LMM can be roughly distributed into five categories: Geometry-based methods assume the presence of at least one pure pixel per endmember [14]–[17]. Statistics-based methods formulate unmixing as a statistical inference problem [18]–[20]. Nonnegative matrix factorization (NMF)-based methods factorize data matrix into a nonnegative mixing matrix and also a nonnegative abundance matrix [21]–[24]. Sparse-based methods amount to finding the optimal subset of signatures in a spectral library [25]–[29]. Deep learning-based methods develop various potential networks to extract abundance and endmember [30]–[37]. Part of the feasible networks are auto-encoder models, which can extract low-dimensional embedding as abundance and supply the weight of decoder as endmember [38]–[40]. For example, EGU-Net considers the properties of endmembers by employing a two-stream siamesed auto-encoder network [41]. By learning two cascaded auto-encoders, CyCU-Net is provided to enhance the unmixing performance [42]. AAENet proposes a novel technique network based on the adversarial auto-encoder [43]. Auto-encoder methods can solve unmixing problem in an simply unsupervised way by means of image reconstruction, thus is also adopted in this paper.

One of the shortcomings in unmixing methods is that the identification of pure endmembers remains constrained by the low spatial resolution. An appropriate guidance, such as super-resolution (SR) technique, may provide a feasible solution to tackle this constraint by improving spatial resolution and sharpening the mixed pixels [44]–[47]. Furthermore, unmixing and super-resolution have been combined by some previous researches together to share both spatial and spectral information. For example, [48] develops a method which jointly solves SR and unmixing to improve SR. A coupled unmixing network with cross-attention for SR is proposed in [49]. [50] combines unmixing and adds smoothing constraints to improve SR. Inspired by the ability of SR to improve spatial resolution and researches mentioned above, SR is considered as an auxiliary task to guide unmixing.

Despite being able to finely combine unmixing and SR, redesigning an appropriate framework remains essential as we aim to fully leverage SR for guiding unmixing. Feasibility of the proposed framework depends on task affinity, the lack

The work was supported by the National Key Research and Development Program of China under Grant 2022YFA1003800, and the Fundamental Research Funds for the Central Universities under grant 63243074. (Corresponding author: Bin Pan)

Ruiying Li, Bin Pan (corresponding author) and Qiaoying Qu are with the School of Statistics and Data Science, KLMDASR, LEBPS, and LPMC, Nankai University, Tianjin 300071, China. (e-mail: liruiying@mail.nankai.edu.cn; panbin@nankai.edu.cn; quqiaoying@mail.nankai.edu.cn).

Xia Xu is with the School of Computer Science and Technology, Tiangong University, Tianjin 300387, China (e-mail: xuxia@tiangong.edu.cn).

Zhenwei Shi is with the Image Processing Center, School of Astronautics, and the State Key Laboratory of Virtual Reality Technology and Systems, Beihang University, Beijing 100191, China (e-mail: shizhenwei@buaa.edu.cn).

[1]Code is released at https://github.com/Lab-PANbin

of which may lead to task gradient conflicts between SR and unmixing. Moreover, an inappropriate structure could result in a trade-off between tasks, thus failing to achieve convergence of the main unmixing task. As a consequence, if we design a framework to guide unmixing by developing SR, two problems are inevitable:

- Task affinity between SR and unmixing is not verified, which leads to task gradient conflicts and casts doubts on whether SR can positively guide unmixing.
- Convergence of unmixing is not guaranteed, which leads to trade-off between tasks and casts doubts on whether unmixing can achieve optimal solution.

In this paper, we propose a **S**R guided **M**ult**I**task **LE**arning method for unmixing (SMILE), which consists of a multitask learning framework and theoretical support. SMILE can generalize positive information from super-resolution to unmixing by learning both shared and specific representations. To validate feasibility of the proposed framework and verify the task affinity, we provide relationship and existence theorems by proving the positive guidance of super-resolution. To guarantee the convergence, we provide the accessibility theorem, which indicates that the optimal solution can be achieved through scalarization. Moreover, for the sake of generalization, universalization and standardization, theorems are backed up with a simplified version of our framework. The main contributions of SMILE can be summarized as follows:

1) We propose a relationship theorem and an existence theorem to verify the positive guidance of super-resolution by proving the task affinity.
2) We propose a new multitask learning framework for unmixing under the guidance of super-resolution, building upon the relationship theorem and existence theorem.
3) We propose an accessibility theorem to verify the convergence of unmixing by proving the accessibility of optimal solution.

The remainder of this paper is structured as follows. Section II states the related work. Section III establishes the proposed theoretical foundation, which proposes a relationship theorem and an existence theorem. Based on the above theoretical analysis, Section IV introduces the proposed algorithm and realizes convergence. Section V describes the experimental results on both synthetic datasets and real datasets.

## II. Related Work

### A. Hyperspectral Unmixing

Hyperspectral unmixing aims to divide observed image $Y \in R^{H \times W \times C}$ into endmembers $E \in R^{p \times C}$ and their corresponding abundance $A \in R^{H \times W \times p}$. Where H and W denote height and width of pixels in $Y$, C describes the count of channels, p is viewed as the number of endmembers [8]. On the basis of LMM, their relationships can be established as:

$$Y = AE + N \qquad (1)$$

where $N$ indicates the noise of image. Taking physical meaning into account, $A$ is subjected to the abundance nonnegative constraint (ANC) and abundance sum-to-one constraint (ASC), which can be symbolized as $A \geq 0$ and $A1_p = 1_{H \times W}$. The core objectives of unsupervised unmixing methods are estimating optimal endmembers and abundance from observed HSIs, which can be exemplified by:

$$\hat{A}, \hat{E} = \underset{A,E}{argmin} \|Y - AE\|^2 \qquad (2)$$
$$s.t. A \geq 0, A1_p = 1_{H \times W}$$

### B. Super–resolution

Super-resolution is to generate a high resolution image $Y_{HR}$ based on a known low resolution image $Y_{LR}$, which can be regarded as one of the inverse problems. There are many types of SR methods for HSIs, including spatial SR algorithms that are designed to improve spatial resolution and recover the estimated high quality image $\hat{Y}_{HR}$. Generally, $Y_{HR}$ and $Y_{LR}$ appear as pairs, then SR process on this situation can be symbolized as:

$$\hat{Y}_{HR} = \underset{\hat{Y}_{HR}}{argmin} \left\| Y_{HR} - \hat{Y}_{HR} \right\|^2 + R(\hat{Y}_{HR}) \qquad (3)$$
$$\hat{Y}_{HR} = f(Y_{LR})$$

where $Y_{HR}$ is adopted as supervised information, $f(\cdot)$ is a SR model and $R(\cdot)$ is a regularizer on the basis of some priors. But these data pairs are difficult to achieve and standard handcrafted priors are hard to build. Established on these, this paper consider blind SR as an auxiliary task for guiding unmixing in an unsupervised way. By adopting a downsampling operator $D(\cdot)$, blind SR can be expressed as:

$$\hat{Y}_{HR} = \underset{\hat{Y}_{HR}}{argmin} \left\| Y_{LR} - D(\hat{Y}_{HR}) \right\|^2 + R(Y_{LR}) \qquad (4)$$
$$\hat{Y}_{HR} = f(Y_{LR})$$

### C. Multitask Learning

Multiple related tasks can be addressed together in multitask learning (MTL) framework, in this way, MTL can transmit shared features and information from one task to another [51]–[56]. Because of the tight relationship between related tasks and the sharing mechanism, main task can be improved by adopting auxiliary task [57]–[61]. Therefore, demonstrating the affinity between two tasks is an important part in MTL. We establish task affinity by following the work in [51]:

*Definition 1:* Given $K$ tasks, define the loss function of task $i$ at time-step $t$ as $\ell_i(y_i, \theta_E^t, \theta_i^t)$, which is parameterized by $\{\theta_E^t\} \cup \{\theta_i^t\}$. $y_i$ is the input of task $i$, $\theta_E^t$ represents the shared parameter and $\theta_i^t$ denotes the specific parameter of task $i$ where $i \in \{1, 2, ..., K\}$. At time-step $t+1$, $\theta_{E|i}^{t+1}$ is defined as the updated shared parameter after a gradient step with respect to task $i$, which can be formulated as:

$$\theta_{E|i}^{t+1} := \theta_E^t - \eta \nabla_{\theta_E^t} \ell_i(y_i, \theta_E^t, \theta_i^t) \qquad (5)$$

*Definition 2:* Based on *Definition 1*, we can define the *task affinity* of task $i$ on task $j$ at time-step $t$ as:

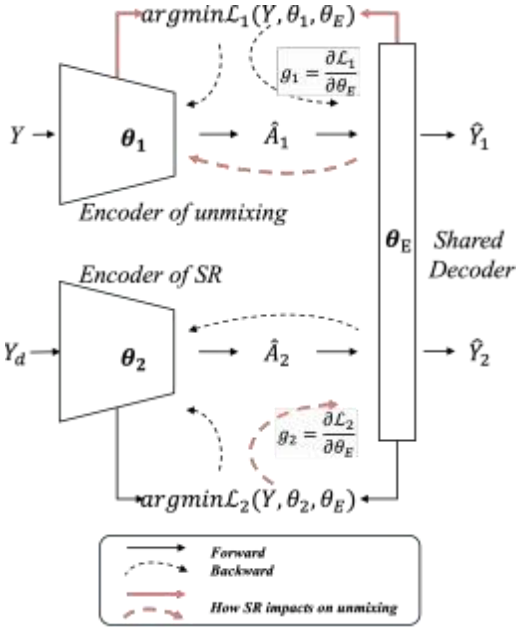

Fig. 1. A generalized model of MTL framework. A shared decoder represents the same endmembers, the specific encoders corresponding to SR and unmixing respectively. It can be regarded as a more generalized, universal and standard version of proposed method.

$$\Lambda_{i \to j}^t := 1 - \frac{\ell_j(y_j, \theta_{E|i}^{t+1}, \theta_j^t)}{\ell_j(y_j, \theta_E^t, \theta_j^t)} \qquad (6)$$

It should be noted that a higher value of $\Lambda_{i \to j}^t$ indicates a stronger *task affinity* of task $i$ on task $j$, which means task $i$ and task $j$ are more related. For example, we regard unmixing as task 1 and SR as task 2, then the guidance of SR to unmixing
can be established by computing $\Lambda_{2 \to 1}^t$.

## III. PROPOSED THEORETICAL FOUNDATION

This section provides theoretical foundation for MTL strategy by describing relationship theorem and existence theorem. First, the feasibility of adopting an MTL framework should be verified. Therefore, theorem 1 (Relationship) demonstrates a high affinity between SR and unmixing. Second, negative transfer in an MTL framework should be avoided. Therefore, theorem 2 (Existence) presents there exists an optimal solution, aligning with the positive guidance of SR on unmixing as mentioned in Remark 2.

The following two theorems demonstrate that the structure of MTL can improve performance of unmixing, providing a theoretical foundation and support for Section IV.

### A. Problem Preliminary

As traditional hard MTL framework, we develop two task-specific parts to deal with SR and unmixing, and one shared decoder to estimate endmembers together. Then SR can guide unmixing and achieve information generalization. For the sake of generalization, universalization and standardization, we establish theorems backed up with a simplified version (Fig.1).The detailed novelty model in Section IV is developed based on this structure.

$$\hat{A}_1 = \text{Encoder}_{\theta_1}^1(Y) \qquad (7)$$
$$\hat{Y}_1 = \hat{A}_1 \theta_E \qquad (8)$$
$$\hat{Y}_{HR} = \text{Encoder}_{\theta_2}^2(Y)\theta_E \qquad (9)$$
$$\hat{Y}_2 = \text{downsampling}(\hat{Y}_{HR}) \qquad (10)$$

where $Y$ is observed hyperspectral image, $\hat{A}_1$ and $\theta_E$ indicate estimated abundance and estimated endmembers respectively. Parameters of two task-specific encoders are $\theta_1$ and $\theta_2$. Weights of one shared decoder actually is predicted endmemers, which can be expressed as $\theta_E$. downsampling($\cdot$) can be considered as a normal operation to reduce spatial resolution of images. $\hat{Y}_1$ and $\hat{Y}_2$ are reconstructed images. The generalized model is shown in Fig.1.

Thus, we develop the loss function of SR and unmixing as $\ell_1$ and $\ell_2$, which can be formulated as:

$$\ell_1(\theta_1, \theta_E) = \left\| Y - \hat{Y}_1 \right\|^2 \qquad (11)$$
$$\ell_2(\theta_2, \theta_E) = \left\| Y - \hat{Y}_2 \right\|^2 \qquad (12)$$

### B. Relationship Between SR and Unmixing

Task affinity, as a theoretical basis, should be verified before adopting MTL. Thus this subsection aims to prove a high affinity between SR and unmixing by demonstrating *Theorem 1*. Moreover, by inversely considering the implications of *Theorem 1*, we also obtain *Remark 1*, which demonstrates there is no gradient conflict between SR and unmixing.

We denote gradient $G_1 = \frac{\partial \ell_1}{\partial(\theta_1, \theta_E)}$, $G_2 = \frac{\partial \ell_2}{\partial(\theta_2, \theta_E)}$. Based on the definition of *task affinity* $\Lambda_{i \to j}^t$, we give the theorem:

*Theorem 1 (Relationship):* The *task affinity* of task 1 unmixing and task 2 SR is bounded by a minimum, that is:

$$\Lambda_{2 \to 1}^t \geq \frac{\eta |G_1|^2}{\ell_1(\theta_E, \theta_1)} \qquad (13)$$

According to the definition of *task affinity* and the descrip- tions of gradient $G_1$ and $G_1$, Eq.(13) equals to :

$$\eta |G_1|^2 + \eta |G_1||G_2|\cos\langle G_1, G_2\rangle \geq \eta |G_1|^2 \qquad (14)$$

by eliminating the same terms on both sides of the equation, we get:

$$\begin{aligned} G_1 \cdot G_2 &\geq 0 \\ \cos\langle G_1, G_2 \rangle &\geq 0 \end{aligned} \qquad (15)$$

where $a \cdot b$ denotes the inner product of two vectors $a$ and $b$. From a geometric point of view, this equals to: the angle between $G_1$ and $G_2$ is less than $90°$.

In order to validate Eq.(14), we design an experiment based on the structure shown in Fig.1. More specifically, we use same synthetic dataset as Section V and we calculate $G_1 \cdot G_2$ in every iteration. Finally, the value falls into the interval $G_1 \cdot G_2 \in [0.81, 0.87]$, that is $0.84 \pm 0.03$. Based on this result, it is obvious that $G_1 \cdot G_2 \geq 0$ is true.

*Remark 1:*

1) *High Task Relatedness*. A high task affinity can be viewed as the solid basement for adopting MTL framework. Theorem 1 allows us to perform analysis in the relationship between SR and unmixing, from which we can solve these two tasks at the same time by developing an MTL framework without any burden.

2) *No Gradient Conflicts*. Gradient conflicts are often cited as one of the reasons why MTL is less effective than single task. Theorem 1 calculates the gradient directions $G_1$ and $G_2$ with their angle so that gradient conflicts turned out not to exist.

### C. Existence of Optimal Solution

In order to avoid negative transfer, this subsection demonstrates the positive guidance of SR on improving unmixing. By demonstrating *Theorem 2*, we prove the existence of optimal solution. By inverting the equivalence of *Theorem 2*, we also obtain *Remark 2*, which demonstrates the positive guidance of SR on improving unmixing.

*Theorem 2 (Existence):* Under the guidance of SR, there exists an optimal solution, which is better than unmixing single task.

Updating the parameters of unmixing can be formulated as:

$$\begin{aligned}\theta^1 &= (\theta_1, \theta_E) - \eta(G_1 + G_2) \\ &= (\theta_1 - \eta G_1, \theta_E - \eta(G_1 + G_2)) \\ \theta^2 &= (\theta_1, \theta_E) - \eta G_1 \\ &= (\theta_1 - \eta G_1, \theta_E - \eta G_1)\end{aligned} \quad (16)$$

where $\theta^1$ is updated in an MTL way, $\theta^2$ is updated in a single task way.

Then, *Theorem 2* is expressed as: $\forall \theta^2$, $\exists \theta^1$ satisfies the following formulation:

$$\ell_1(\theta^1) \leq \ell_1(\theta^2) \quad (17)$$

Let $\theta^0 = (\theta_1, \theta_E)$ be initial parameters of loss $\ell_1$. There are two different updating methods as described in Eq.(16). We first consider about taking a gradient step on the parameters only using the gradient of task 1, which means we get $\theta^2$.

By using Taylor expansion, that is:

$$\begin{aligned}\ell_1(\theta^2) &= \ell_1(\theta^0 - \eta G_1) \\ &\geq \ell_1(\theta^0) + \frac{\partial \ell_1}{\partial \theta^0}(-\eta G_1) \\ &= \ell_1(\theta^0) - \eta |G_1|^2\end{aligned} \quad (18)$$

Under *Theorem 1*:

$$\Lambda_{2 \to 1}^t = 1 - \frac{\ell_1(\theta^0 - \eta(G_1 + G_2))}{\ell_1(\theta^0)} \geq \frac{\eta |G_1|^2}{\ell_1(\theta^0)} \quad (19)$$

equals to:

$$\ell_1(\theta^0) - \ell_1(\theta^0 - \eta(G_1 + G_2)) \geq \eta |G_1|^2 \quad (20)$$

based on the above:

$$\begin{aligned}\ell_1(\theta^1) &= \ell_1(\theta^0 - \eta(G_1 + G_2)) \\ &\leq \ell_1(\theta^0) - \eta |G_1|^2 \leq \ell_1(\theta^2)\end{aligned} \quad (21)$$

This finishes the proof of Theorem 2.

*Remark 2:*

1) *Positive Guidance*. From Theorem 2, taking a gradient step on the parameters using the combined gradient of SR and unmixing, that is $G_1 + G_2$, reduces $\ell_1$ more than taking a gradient step on the parameters using the gradient of unmixing, that is $G_1$. This conclusion indicates the positive guidance of SR on unmixing.

From the above three aspects in Remark 1 and 2, we consider that SR can indeed improve the result of unmixing because the two tasks have high task affinity, with no gradient conflicts and MTL reduces more loss than unmixing single task. On the basis of above theorems, we propose novelty MTL method for unmixing in Section IV.

## IV. PROPOSED METHOD

In Section III, a detailed theoretical foundation is presented, which explores that MTL structure can improve the performance of unmixing. On the basis of the analysis in Section III, in this section, we will introduce the proposed SMILE method. SMILE framework is composed of task-specific part (section IV-B) and task-shared part (section IV-C). Moreover, to guarantee the convergence of the main task in MTL, we provide accessibility theorem (section IV-D) and for further improving the theoretical analysis.

### A. Multitask Learning Framework of SMILE

SMILE takes unmixing as a main task and develops SR as an auxiliary task to improve unmixing. We adopt this general assumption that low resolution image Y and estimated high resolution image $\hat{Y}_{HR}$ have same endmember and different abundances. As a result, the weight $\theta_E$ of shared decoder $g_E$ represents the shared endmember. On the basis of LMM, latent feature map $\hat{A}_1$ and $\hat{A}_{HR}$ are regarded as abundances in Y and $\hat{Y}_{HR}$, respectively.

Following hard MTL structure mentioned in Section III, SMILE is designed as an unsupervised end-to-end framework, which is composed of task-specific part and task-shared part. The task-specific part contains an auto-encoder for unmixing and a deep image prior (DIP) structure for SR, which is described as two brunches of an up-down structure. As described in Fig.2, the encoder in upper unmixing brunch takes observed Y as input and outputs abundance $\hat{A}_1$, which is considered as input of decoder to get reconstructed image $\hat{Y}_1$. The lower SR brunch first generates random noise $l_Y$ and $l_k$, which are inputs Input from two different directions. From left to right, $l_Y$ is fed into DIP structure to get high resolution abundance $\hat{A}_{HR}$ and high resolution image $\hat{Y}_{HR}$. From right to left, $l_k$ is is processed by a generator to get downsampling operator $\hat{k}$. By combining $\hat{Y}_{HR}$ and $\hat{k}$, DIP structure outputs another reconstructed image $\hat{Y}_2$.

In Section III, relationship theorem (Theorem 1) is provided by analyzing gradients $g_1$ and $g_2$, existence theorem (Theorem 2) is verified by computing $L_1$. Moreover, as described in Fig.2 the convergence of SMILE is guaranteed by accessibility theorem (Theorem 3), which validates that the optimal solution $\hat{A}_1$ of unmixing can be achieved by scalarization way.

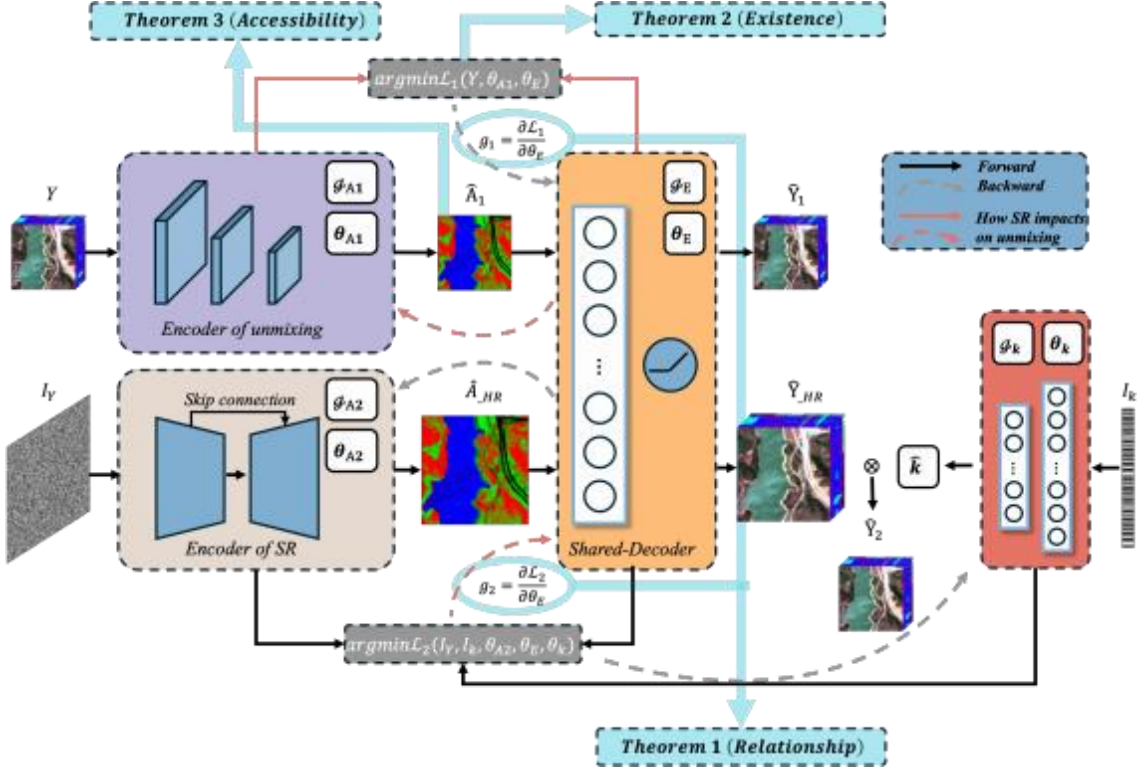

Fig. 2. Overview of the proposed SMILE framework. There exists two subnetworks with specific encoders and shared decoder. Gradients transfer along the direction of the arrows, thus making information generalization.

### B. Task–specific Part

The positive guidance of SR for unmixing is realized by multi-task learning framework, which consists of the task-specific parts and the task-shared part. In which task-specific part is consists of an auto-encoder for unmixing and a deep image prior (DIP) structure for SR. The two tasks are described as two brunches of an up-down structure.

*1) Autoencoder for Hyperspectral Unmixing:* We now introduce how to solve unmixing problem by developing an auto-encoder. The objective function is expressed as Eq.(2), in which abundance and endmembers are estimated by minimizing the reconstruction error under the constraints of ANC and ASC. This structure is composed of an encoder $g_{A1}$ with parameter $\theta_{A1}$ and a decoder $g_E$ with parameter $\theta_E$.

Observed image $Y$ is input and abundance $\hat{A}_1$ is a low dimensional representation obtained as the output of encoder $g_{A1}$. Then $\hat{A}_1$ is viewed as the input of $g_E$ and reconstructed image $\hat{Y}_1$ is obtained. These above process can be expressed as:

$$\hat{A}_1 = g_{A1}(Y) \tag{22}$$
$$\hat{Y}_1 = g_E(\hat{A}_1) \tag{23}$$

Naturally, unmixing problem considering reconstruction error in this part can be reformulated as:

$$\hat{\theta}_{A1}, \hat{\theta}_E = \underset{\hat{\theta}_{A1},\hat{\theta}_E}{\arg\min} L_1(Y, \theta_{A1}, \theta_E) = \underset{\hat{\theta}_{A1},\hat{\theta}_E}{argmin} \left\| Y - \hat{Y}_1 \right\|^2 \tag{24}$$

where the weight matrix $\theta_E$ connecting hidden layer and output layer is endmember actually.

*2) Deep Image Prior for Super–resolution:* In this part, we estimate the latent high resolution image and the unknown downsampling operator by developing DIP structure.

Formulated by Eq.(3), super-resolution generates high spatial resolution image from a known low resolution input. Due to the absence of corresponding high spatial resolution image $Y_{HR}$, there is no ground truth available to supervise the generation process. Thus SR in this paper as an auxiliary task can only be developed in an unsupervised way.

Rather than using handcrafted image priors, DIP-based SR employs deep network structures to capture image priors. In this way, DIP can achieve high spatial resolution image without label and provide prior for image generation, which are typically presented as:

$$\hat{\theta} = \underset{\theta}{\arg\min} \| Y_{LR} - f_\theta(z) \otimes K \|^2$$
$$Y_{HR} = f_{\hat{\theta}}(z) \tag{25}$$

where $z$ is a randomly-initialized 3D tensor with the same shape as $Y_{HR}$. $K$ is a downsample operator. $f_\theta(z)$ can be considered as an image generator, which can recover high resolution image from noise.

More specifically, to better guide and improve unmixing problem, we divide high resolution images into endmembers and their corresponding abundance $\hat{A}_{HR}$, with the downsam-pling operator being learnable. As described in Fig.2, the above process can be expressed as:

$$\hat{A}_{HR} = g_{A2}(I_Y) \tag{26}$$
$$\hat{Y}_{HR} = g_E(\hat{A}_{HR}) \tag{27}$$
$$\hat{Y}_2 = \hat{Y}_{HR} \otimes g_k(I_k) \tag{28}$$

where $I_Y$ and $I_k$ are random Gaussian noise inputs. Generator $g_{A2}$ with parameter $\theta_{A2}$ converts $I_Y$ to the predicted high resolution abundance. $\hat{A}_{HR}$ is subsequently reconstructed to $\hat{Y}_{HR}$ weighted by $\theta_E$. Generator $g_k$ with parameter $\theta_k$ outputs $\hat{K}$ that denotes the estimated downsampling operator. Based on these, SR problem in this paper as an auxiliary task can be reformulated as:

$$\begin{aligned}\hat{\theta}_{A2},\hat{\theta}_E,\hat{\theta}_k &= \underset{\hat{\theta}_{A2},\hat{\theta}_E,\hat{\theta}_k}{argmin}\mathcal{L}_2(I_Y,I_k,\theta_{A2},\theta_E,\theta_k) \\ &= \underset{\hat{\theta}_{A2},\hat{\theta}_E,\hat{\theta}_k}{argmin}\left\|Y-\hat{Y}_2\right\|^2\end{aligned} \tag{29}$$

### C. Task-shared Part

Based on LMM, both low-resolution image $\hat{Y}_1$ and high-resolution image $\hat{Y}_{HR}$ can be divided into endmembers $\theta_E$ and their corresponding abundances $\hat{A}_1$ and $\hat{A}_{HR}$. Among them, $\theta_E$ is the same in $\hat{Y}_1$ and $\hat{Y}_{HR}$, $\hat{A}_1$ can be regarded as spatially transformation about $\hat{A}_{HR}$. $\hat{Y}_2$ is reconstructed by $\hat{Y}_{HR}$ and a learnable downsampling operator. Estimated $\hat{Y}_1$ and $\hat{Y}_2$ can be expressed as:

$$\hat{Y}_1 = g_{A1}(Y)\theta_E \tag{30}$$
$$\hat{Y}_2 = g_{A2}(I_Y)\theta_E \otimes g_k(I_k) \tag{31}$$

Therefore, the task-shared part is designed to update endmembers $\theta_E$. The shared-decoder $g_E$ is optimized by both unmixing specific part and SR specific part. This MTL framework with shared part can generalize information from one task to another and provide more detailed guidance on unmixing.

From the above discussions, the proposed network consists of two brunches: unmixing and SR. This framework can generate main five estimations: reconstructed image $\hat{Y}_1$ with its abundance $\hat{A}_1$, predicted high spatial resolution image $\hat{Y}_{HR}$ with its abundance $\hat{A}_{HR}$, same endmember $\theta_E$ in $\hat{Y}_1$ and $\hat{Y}_{HR}$. Corresponding to the two brunches, we propose $L_1$ for unmixing and $L_2$ for SR. The two loss functions are described as:

$$L_1(Y,\theta_{A1},\theta_E) = \|Y - g_{A1}(Y)\theta_E\|^2 \tag{32}$$
$$L_2(I_Y,I_k,\theta_{A2},\theta_E,\theta_k) = \|Y - g_{A2}(I_Y)\theta_E \otimes g_k(I_k)\|^2 \tag{33}$$

---

**Algorithm 1**: SMILE

**Input:** Hyperspectral image $Y$, num-iter: T
**Output:** endmember $\theta_E$, abundance $\hat{A}_1$
**Estimation:** $\hat{Y}_1$, $\hat{Y}_{HR}$, $\hat{A}_{HR}$, $\hat{K}$
**Initialization:**
1: Sample $I_Y$ and $I_k$ from standard normal distribution,
2: Random initialization for $\theta_{A1}^0$, $\theta_{A2}^0$, $\theta_k^0$
3: Initializing $\theta_E^0$ by VCA [14]
**while** t < T **do:**
  Calculate $\hat{A}_1^t = g_{A1}^t(Y)$, $\hat{Y}_1^t = g_E^t(\hat{A}_1)$
  $\hat{A}_{HR}^t = g_{A2}^t(I_Y)$, $\hat{Y}_{HR}^t = g_E^t(\hat{A}_{HR})$, $\hat{K}^t = g_k^t(I_k)$
  Compute the gradient $G_{A1}^t = \frac{\partial\mathcal{L}}{\partial\theta_{A1}^t}$, $G_E^t = \frac{\partial\mathcal{L}}{\partial\theta_E^t}$
  $G_{A2}^t = \frac{\partial\mathcal{L}}{\partial\theta_{A2}^t}$, $G_k^t = \frac{\partial\mathcal{L}}{\partial\theta_k^t}$
  Update $\theta_{A1}^{t+1}$, $\theta_{A2}^{t+1}$, $\theta_k^{t+1}$ and $\theta_E^{t+1}$
  t = t + 1
**end for**
abundance $\hat{A}_1 = g_{A1}^T(Y)$ and endmemer $\theta_E^T$

---

Moreover, to guarantee the constraints in Eq.(2) and sparsity, we limit the sum of estimated abundance $\hat{A}_1$ to satisfy these constraints, which is given by

$$\mathcal{L}_3(\hat{A}_1) = \left\|\hat{A}_1\right\|_* = tr(\sqrt{\hat{A}_1^T\hat{A}_1}) \tag{34}$$

$$\mathcal{L}_4(\hat{A}_1) = \sum_{i=1}^{H\times W}\left\|1 - \sum_{j=1}^{p}\hat{a}_{ji}\right\|_1 \tag{35}$$

where $\hat{A}_1$ is reshaped as $\hat{A}_1 \in R^{(H\times W)\times p}$. Then the total loss function is obtained by linear scalarization, which combines all loss functions by a weighted sum:

$$\mathcal{L} = \sum_{i=1}^{4}\alpha_i\mathcal{L}_i \tag{36}$$

here, the convex coefficients $\|\alpha_i\|$ satisfies:

$$\alpha_i \geq 0 \quad and \quad \sum_{i=1}^{4}\alpha_i = 1 \tag{37}$$

and are typically obtained via grid search.

The generalization between unmixing and SR can be realized through gradient backward propagation and linear scalarization, which are common method for jointly optimizing objective functions. As summarized in **Algorithm1**, this approach enables the estimation of images and abundances at higher spatial resolutions, thus improving the results of unmixing.

### D. Accessibility of Optimal Solution

On the basis of *Theorem 1* and *Theorem 2*, this subsection proves the convergence of unmixing by demonstrating *Theo-rem 3*, which validates that the optimal solution of unmixing can be achieved by scalarization way.

*Lemma 1:* MTL framework optimized by linear scalarization is capable of fully exploring optimal solution, if it satisfies the following conditions:

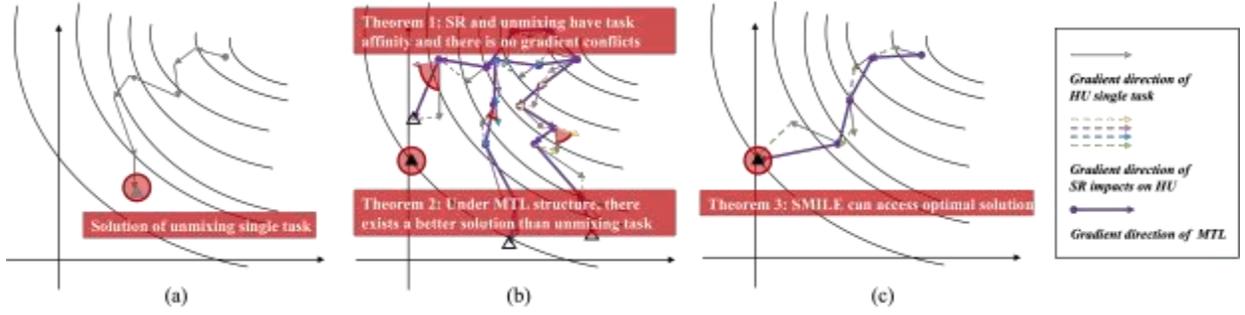

Fig. 3. Optimization process of different methods. Fig.3(a) represents the original gradient directions of unmixing single task. Fig.3(b) describes different effects of SR working on unmixing, which may not be able to obtain the optimal solution. Fig.3(c) displays that SMILE can reduce loss function of unmixing more than unmixing single task, and achieve optimal.

1) Network should be over-parameterization.
2) There exists no gradient conflicts in different tasks.
3) Coefficients of loss functions weights should be convex.

*Lemma 1* is a restate of works in [53]. More detailed proof can be found in this paper.

*Theorem 3 (Accessibility):* SMILE is capable of fully ex - ploring optimal solution.

We will validate the conditions of *Lemma 1* item by item to prove the conclusion.

1) We denote the number of tasks as k and the account of network as q. It is obvious that k = 2 and q ≥ 4. Then, we get:

$$q \gg k \tag{38}$$

that is over-parameterization.
2) This condition is mentioned in *Remark 1*.
3) This condition is satisfied in Eq.(37).

To illustrate the effectiveness of all the theorems, we analyze their roles in the optimization process, as displayed in Fig.3. Solutions (triangles) closer to the lower-left contour are considered better. In Fig.3(a), the gray solid lines represent the original gradient directions of the single unmixing task. In Fig.3(b) and Fig.3(c), the dashed lines of different colors denote the influence of SR task on the gradient directions of unmixing task. The thicker purple solid lines describe the final gradient directions of unmixing in MTL framework.

Fig.3(a) describes the original gradient directions of unmixing single task. Fig.3(b) displays different effects of SR working on unmixing. It is worthy to note that the angles between gradient directions are less than $90°$, which has been demonstrated in *Theorem 1*. Fig.3(b) also expresses that the existence of optimal solution (solid triangle in Fig.3(b)) is guaranteed by *Theorem 2*, which verifies that MTL structure can achieve a superior solution compared to the single unmix- ing task.

In Fig.3(c), the solid triangle can be realized by purple gradient directions. This description is supported by *Theorem 3*, which shows that SMILE can reduce loss function of unmixing more than unmixing single task, and achieve the better solution in scalarization way.

## V. EXPERIMENTS

In this section, we validate the effectiveness of the proposed method SMILE. We conduct comparison experiments on both synthetic and real datasets with seven different unmixing methods. We also conduct ablation study on a real dataset.

### A. Hyperspectral Data Description

We generate two synthetic datasets and adopt two real datasets to demonstrate the performance of various unmixing algorithms.

*1) Synthetic Datasets:* The first synthetic dataset is constructed by endmember selection and abundance simulation. Observed HSIs $Y \in R^{N \times C}$ contains C spectral bands and N pixels, abundance $A \in R^{N \times p}$ follows a Dirichlet distribution and endmembers $E \in R^{p \times C}$ are chosen from a spectral library. Image size N is fixed at $64 \times 64$, the number of endmember p is selected from 3 to 10 and C equals to 224 means that Y contains 224 spectral bands. The signal-to-noise ratios (SNR) of first synthetic dataset are 20-/30-/40-dB, respectively. SNR is defined as follow:

$$SNR = 10 \times log\left(\frac{E(\|Y\|_F^2)}{E(\|N\|_F^2)}\right) \tag{39}$$

The second synthetic dataset is generated by following literature [29] and SNR is set as 30-dB. Image size N is fixed at $100 \times 100$, the number of endmember p is selected as 4.

*2) Real Datasets:* Samson contains $95 \times 95$ pixels. Each pixel is recorded at 156 channels. There are three targets: soil, tree and water. Jasper Ridge contains $100 \times 100$ pixels in it. Each pixel is recorded at 198 channels. There are four endmembers latent in this data: road, soil, water and tree.

### B. Experimental Setup

*1) Performance Metrics:* We consider three different metrics in this paper to evaluate the performance. Root mean square error (RMSE) is calculated between estimated abundance and real abundance, which is given by

$$RMSE = \sqrt{\frac{1}{N}\sum_{i=1}^{N}\|a_i - \hat{a}_i\|_2^2} \tag{40}$$

TABLE I
RMSE Result of SMILE, CyCU-Net, EGU-Net, BUDDIP, UnDIP, GLA, PGMSU and DAEN on synthetic data 1

| SNR | p | SMILE | CyCU-Net | EGU-Net | BUDDIP | UnDIP | GLA | PGMSU | DAEN |
|---|---|---|---|---|---|---|---|---|---|
| 20-dB | 3 | 0.3121 | 0.3924 | 0.2712 | 0.2981 | **0.2822** | 0.3136 | 0.4269 | 0.4661 |
|  | 4 | 0.3571 | 0.3648 | 0.3475 | **0.3004** | 0.3104 | 0.3234 | 0.3459 | 0.4929 |
|  | 5 | **0.2470** | 0.3963 | 0.3515 | 0.2687 | 0.2665 | 0.3179 | 0.3673 | 0.3347 |
|  | 6 | **0.2508** | 0.3482 | 0.4162 | 0.2559 | 0.2559 | 0.2939 | 0.3325 | 0.3865 |
|  | 7 | 0.2469 | 0.2713 | 0.5549 | 0.2339 | **0.2334** | 0.2631 | 0.2952 | 0.3252 |
|  | 8 | **0.2216** | 0.2845 | 0.5326 | 0.2257 | 0.2353 | 0.2421 | 0.2823 | 0.3019 |
|  | 9 | **0.2072** | 0.2952 | 0.4646 | 0.2232 | 0.2235 | 0.2332 | 0.2933 | 0.3041 |
|  | 10 | **0.1962** | 0.2519 | 0.5076 | 0.2021 | 0.2128 | 0.2308 | 0.2583 | 0.2599 |
| 30-dB | 3 | 0.3787 | 0.3840 | 0.3750 | 0.1562 | **0.2961** | 0.3577 | 0.4430 | 0.3763 |
|  | 4 | 0.3481 | 0.3557 | 0.2329 | **0.3169** | 0.2791 | 0.3297 | 0.3919 | 0.4836 |
|  | 5 | **0.2451** | 0.3480 | 0.3785 | 0.2603 | 0.2687 | 0.3109 | 0.3799 | 0.3904 |
|  | 6 | 0.2678 | 0.3588 | 0.5860 | 0.2496 | **0.2535** | 0.2847 | 0.3346 | 0.3718 |
|  | 7 | **0.2452** | 0.2971 | 0.5259 | 0.2493 | 0.2572 | 0.2684 | 0.2966 | 0.3124 |
|  | 8 | **0.2127** | 0.3219 | 0.2651 | 0.2278 | 0.2317 | 0.2376 | 0.2877 | 0.3042 |
|  | 9 | **0.2015** | 0.2679 | 0.4244 | 0.2201 | 0.2129 | 0.2316 | 0.2774 | 0.2432 |
|  | 10 | **0.1947** | 0.2285 | 0.4027 | 0.2071 | 0.2110 | 0.2266 | 0.2516 | 0.2809 |
| 40-dB | 3 | 0.2131 | 0.4761 | **0.2088** | 0.2864 | 0.3307 | 0.3383 | 0.3994 | 0.4927 |
|  | 4 | **0.2400** | 0.5242 | 0.2587 | 0.3003 | 0.3243 | 0.3637 | 0.4092 | 0.4855 |
|  | 5 | **0.2327** | 0.4542 | 0.4118 | 0.2954 | 0.2639 | 0.3037 | 0.3775 | 0.4464 |
|  | 6 | **0.2445** | 0.3720 | 0.4985 | 0.2597 | 0.2589 | 0.2755 | 0.3347 | 0.2820 |
|  | 7 | 0.2383 | 0.3361 | 0.3193 | **0.2371** | 0.2458 | 0.2595 | 0.3079 | 0.3346 |
|  | 8 | **0.2239** | 0.2720 | 0.5231 | 0.2337 | 0.2312 | 0.2454 | 0.2842 | 0.3176 |
|  | 9 | **0.1982** | 0.2608 | 0.3806 | 0.2163 | 0.2195 | 0.2263 | 0.3017 | 0.2849 |
|  | 10 | **0.2005** | 0.2548 | 0.5454 | 0.2051 | 0.2180 | 0.2259 | 0.2529 | 0.2927 |

TABLE II
RMSE, AAD, SAD, SAD1, SAD2, SAD3 and SAD4 Results of SMILE, CyCU-Net, EGU-Net, BUDDIP, UnDIP, GLA, PGMSU and DAEN on synthetic data 2

| Metrics | SMILE | CyCU-Net | EGU-Net | BUDDIP | UnDIP | GLA | PGMSU | DAEN |
|---|---|---|---|---|---|---|---|---|
| RMSE | 0.3989 | 0.2691 | 0.43592 | **0.2689** | 0.3436 | 0.4164 | 0.4338 | 0.4986 |
| AAD | 15.4321 | 11.3027 | 23.2430 | 9.5524 | 17.5240 | 22.8122 | **5.8359** | 24.4784 |
| SAD | **1.5790** | 3.1188 | 19.0318 | 3.6588 | 16.4735 | 22.7096 | 3.6887 | 4.1155 |
| SAD1 | **3.1635** | 6.7239 | 15.7734 | 5.3113 | 23.8791 | 22.8037 | 8.5120 | 5.1625 |
| SAD2 | **0.3510** | 0.9784 | 18.2436 | 1.7564 | 19.0986 | 23.7185 | 0.8579 | 4.0927 |
| SAD3 | 1.0555 | 1.3142 | 22.8683 | 2.4120 | 10.9561 | 25.2356 | **0.7437** | 2.9128 |
| SAD4 | **1.7461** | 3.4586 | 21.2418 | 5.1549 | 11.9600 | 20.0803 | 4.6409 | 4.2940 |

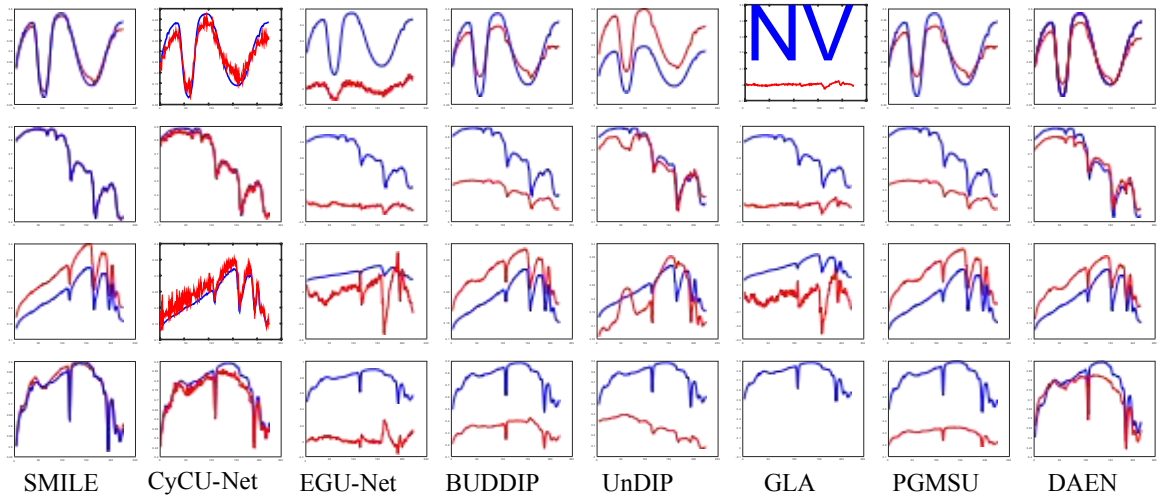

SMILE    CyCU-Net    EGU-Net    BUDDIP    UnDIP    GLA    PGMSU    DAEN

Fig. 4. Extracted endmember comparison from top to bottom corresponding to endmembers number from 1 to 4 on synthetic data 2. These pictures from left to right are respectively obtained by SMILE, CyCU-Net, EGU-Net, BUDDIP, UnDIP, GLA, PGMSU and DAEN with 30-dB noise.

where $a_i$ and $\hat{a}_i$ describe the the ith pixel of predicted and real value. And the abundance angle distance (AAD) is also employed to measure the quality of estimated abundance by the following definition:

$$\text{AAD} = \frac{1}{p} \sum_{i=1}^{p} \arccos \left( \frac{a_i^T \hat{a}_i}{\|a_i\|_2 \|\hat{a}_i\|_2} \right) \quad (41)$$

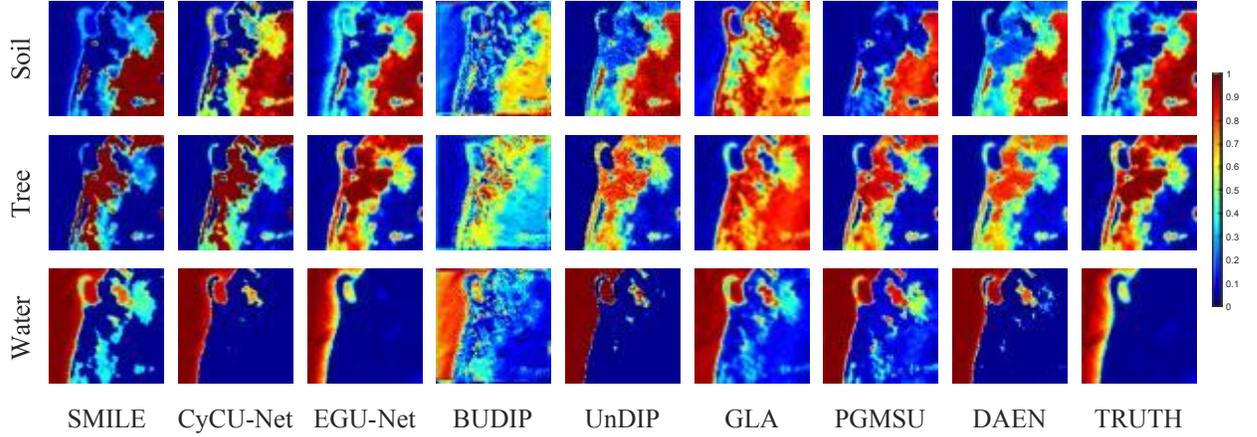

Fig. 5. Comparison of abundance maps on Samson data. (Left to right) Abundance maps obtained by SMILE, CyCU-Net, EGU-Net, BUDDIP, UnDIP, GLA, PGMSU and DAEN respectively. (Top to bottom) Maps corresponding to Soil, Tree, Water.

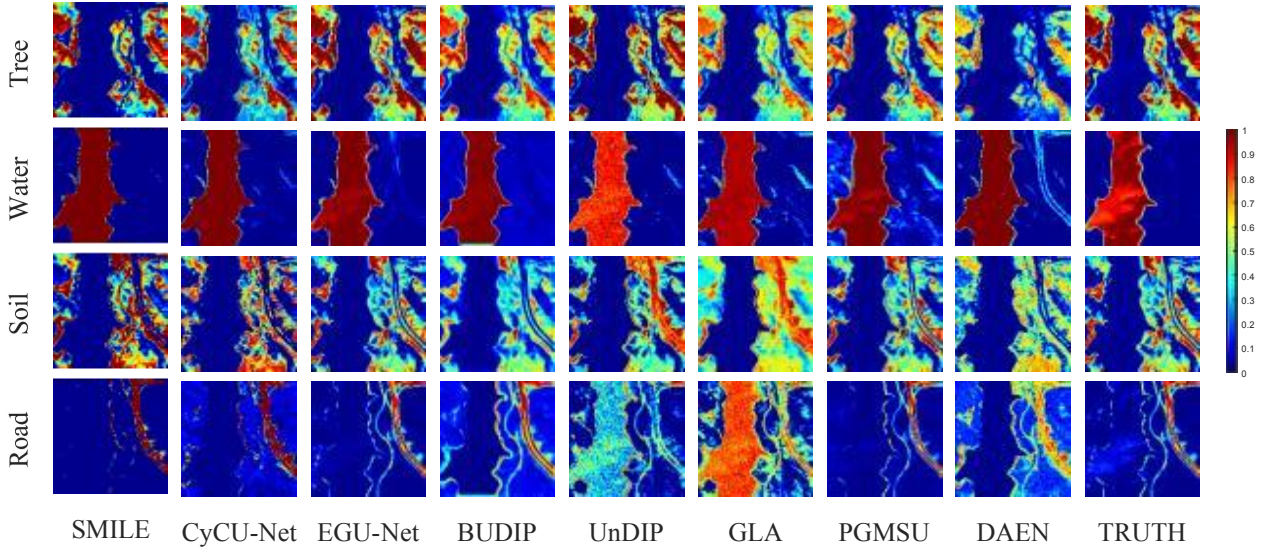

Fig. 6. Comparison of abundance maps on Jasper data. (Left to right) Abundance maps obtained by SMILE, CyCU-Net, EGU-Net, BUDDIP, UnDIP, GLA, PGMSU and DAEN respectively. (Top to bottom) Maps corresponding to Tree, Water, Soil, Road.

similarly, we adopt spectral angle distance (SAD) to evaluate the estimated endmember $e_i$ and the true value $\hat{e}_i$, where SAD of ith endmember is described as :

$$SAD_i = arccos\left(\frac{e_i^T \hat{e}_i}{\|e_i\|_2 \|\hat{e}_i\|_2}\right) \quad (42)$$

as there are p endmembers, $SAD_i$ can be averaged over p to generate a scalar quantities SAD:

$$SAD = \frac{1}{p}\sum_{i=1}^{p} SAD_i \quad (43)$$

*2) Comparison Algorithms:* We compare the proposed method with some state-of-the-art deep learning unmixing methods including cycle-consistency unmixing network (CyCU-Net) [42], endmember-guided unmixing network (EGU-Net) [41], double DIP (BUDDIP) [33], hyperspectral unmixing using deep image prior (UnDIP) [39], global-Local smoothing autoencoder (GLA) [31], probabilistic generative model for hyperspectral unmixing (PGMSU) [34] and deep autoencoder network (DAEN) [38]. CyCU-Net learns two cascaded auto-encoders to enhance the unmixing performance. EGU-Net fully considers the properties of endmembers extracted from the hyperspectral imagery and develops a general deep learning approach for unmixing. BUDDIP is inspired by DIP and proposes an unsupervised framework that can be used for both linear and nonlinear blind unmixing models. UnDIP utilizes a conventional geometrical approach for endmember extraction and a new deep convolutional neural network for abundance estimation. GLA is an unsupervised model which aims at exploring the local homogeneity and the global self-similarity of hyperspectral imagery. PGMSU follows an encoder–decoder network architecture for simultaneously estimating abundances and endmembers. DAEN specifically addresses the presence of outliers in hyperspectral data.

### C. Unmixing Performance

Table I lists the quantitative performance comparison on the synthetic data 1, we set p as the number of endmembers, which varies from 3 to 10. Moreover, the noise is considered from 20-dB to 40-dB. We only compare RMSE in this table to measure different abundances generated by eight algorithms. From Table I, it is indicated that RMSE results of SMILE is less than the other methods for a large proportion, especially with high SNR. In some cases, BUDIP and UnDIP also perform preferable results, which makes sense for DIP to be adopted for unmixing.

Table II lists the quantitative performance comparison on the synthetic data 2 with 30-dB noise. RMSE and AAD are viewed two metrics to quantify predicted abundance, SAD is employed to evaluate the estimated endmembers. In order to analyze endmembers more accurate, we also take $SAD_i$ for ith endmember. From Table II, it is indicated that estimated abundance of SMILE is not the best. But the predicted endmembers perform better in most cases. To illustrate this phenomenon more clearly, we provider qualitative result in Fig.4. In this figure, we describe groundtruth endmembers by drawing blue lines, the predicted endmembers of different methods are presented by drawing red lines. The closer the two lines are and the smaller the SAD value, the better the results.

Fig.5 and Fig.6 present the corresponding abundance maps as the qualitative result in the Samson data set and Jasper data set, respectively. Even though we provide groundtruth abundance as reference, we contend that a stronger contrast between light and dark in the abundance maps indicates better performance, rather than mere proximity to the ground truth image. On the basis of this, it can be determined that proposed SMILE achieves a better performance in terms of the estimated abundance.

### D. Ablation Study

In this part, we perform the ablation analysis to investigate the effectiveness of SR component for unmixing applications. It should be noted that DAEN can be taken as the baseline for deep learning unmixing with auto-encoder. These comparisons have been described in the previous section.

Table III lists the three different metrics results on the real datasets, which indicate the capability of SR guiding unmixing. Fig.7 describes the process of abundance change, which not only compare DAEN and SMILE, but also display the abundance generated by SR. It should be noted that the abundance of the SR branch is not adopted for comparison, but is employed to show its positive effect on unmixing.

TABLE III
QUANTITATIVE RESULTS OF SMILE AND DAEN ON REAL DATASETS

| Dataset | Metrics | SMILE | DAEN |
|---|---|---|---|
| Samson | RMSE | **0.1675** | 0.2131 |
|  | AAD | **17.7118** | 19.3309 |
|  | SAD | 17.9714 | **11.5674** |
| Jasper | RMSE | **0.1313** | 0.2929 |
|  | AAD | **20.1533** | 25.9631 |
|  | SAD | **9.6420** | 10.0961 |

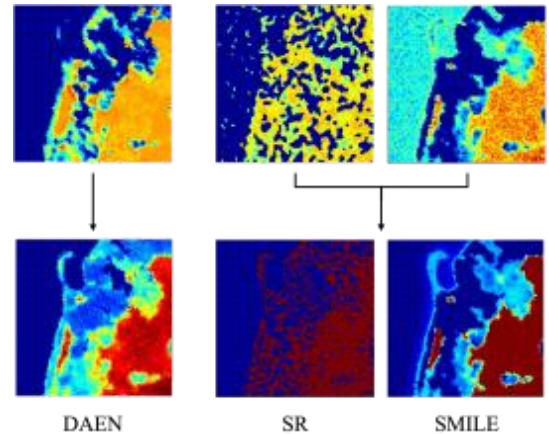

Fig. 7. Qualitative Results of DAEN and SMILE on Samson Dataset

## VI. CONCLUSION

In this paper, we propose a super resolution guided multi-task learning method for hyperspectral unmixing (SMILE),which consists of a newly designed framework and theoretical support. The proposed framework generalizes positive information from super-resolution to unmixing by learning both shared and specific representations. To validate feasibility of the proposed framework and verify task affinity, we provide relationship and existence theorems by proving the positive guidance of super-resolution. To guarantee the convergence, we provide the accessibility theorem by proving the optimal solution of unmixing. The major contributions of SMILE include designing a new framework for unmixing under the guidance of super-resolution, and providing progressive theoretical support. Experiments with synthetic and real hyperspectral datasets validate the effectiveness of SMILE compared with other unmixing approaches.